\documentclass[conference]{IEEEtran}
\IEEEoverridecommandlockouts
\usepackage{cite}
\usepackage{amsmath,amssymb,amsfonts}
\usepackage{algorithmic}
\usepackage{graphicx}
\usepackage{textcomp}
\usepackage{xcolor}
\usepackage{caption}
\usepackage{float}
\usepackage{dblfloatfix}
\usepackage{subcaption}
\usepackage{array}
\usepackage{makecell}
\usepackage{multirow}
\usepackage{tikz}
\usepackage{pgfplots}
\usepackage{soul}
\usepackage{color, colortbl}
\usepackage{url}

\DeclareUnicodeCharacter{2212}{\ensuremath{-}}
\def\BibTeX{{\rm B\kern-.05em{\sc i\kern-.025em b}\kern-.08em
    T\kern-.1667em\lower.7ex\hbox{E}\kern-.125emX}}

\newcommand{\mydiamond}{%
  \sbox0{$\lozenge$}%
  \usebox0\kern-.5\wd0\clap{\raisebox{.1ex}{\scalebox{.7}[1]{$-$}}}\kern.5\wd0%
}

\begin{document}

\title{PolyLUT: Learning Piecewise Polynomials for Ultra-Low Latency FPGA LUT-based Inference\\

\author{\IEEEauthorblockN{Marta Andronic and George A. Constantinides}
\IEEEauthorblockA{Department of Electrical and Electronic Engineering \\
Imperial College London, UK\\
Email: \{marta.andronic18, g.constantinides\}@imperial.ac.uk}
}

}
\maketitle

\begin{abstract}
Field-programmable gate arrays (FPGAs) are widely used to implement deep learning inference. Standard deep neural network inference involves the computation of interleaved linear maps and nonlinear activation functions. Prior work for ultra-low latency implementations has hardcoded the combination of linear maps and nonlinear activations inside FPGA lookup tables (LUTs). Our work is motivated by the idea that the LUTs in an FPGA can be used to implement a much greater variety of functions than this. In this paper, we propose a novel approach to training neural networks for FPGA deployment using {\em multivariate polynomials} as the basic building block. Our method takes advantage of the flexibility offered by the soft logic, hiding the polynomial evaluation inside the LUTs with minimal overhead. We show that by using polynomial building blocks, we can achieve the same accuracy using considerably fewer layers of soft logic than by using linear functions, leading to significant latency and area improvements. We demonstrate the effectiveness of this approach in three tasks: network intrusion detection, jet identification at the CERN Large Hadron Collider, and handwritten digit recognition using the MNIST dataset. 
\end{abstract}

\section{Introduction and Motivation}
Deep learning (DL) is a powerful technique for extracting abstract features from raw data and has shown remarkable success in various applications such as image classification and natural language processing~\cite{goodfellow}. Deploying these models to the cloud leads to the inability to produce results in real-time, security concerns, and increased latency and round-trip delay~\cite{edge}. As a result, there is a growing interest in deploying DL models on edge devices. However, due to the models' high computational cost and power and memory requirements, this has proven to be a challenging task~\cite{esurvey}.

An approach that has been proposed to address the high computational cost of neural networks and the infeasibility of deployment on memory-constrained platforms is the binary neural network (BNN)~\cite{edge}. Through the binarization of weights and activations, on-chip memory is saved and, together with the replacement of multi-bit multipliers with lightweight XNOR gates, it has been suggested that BNNs can provide an area- and energy- efficient solution that removes redundancy while maintaining high classification accuracy after retraining.

However, BNNs are still not perfect for FPGAs because they do not take advantage of the underlying FPGA architecture. In particular, the XNOR gates and parts of the popcount logic end up mapped onto $K$-LUTs, which are capable of implementing arbitrary Boolean functions with limited support. Prior works that use LUTs as more than just XNORs can be divided in two categories: Differentiable LUTs (LUTNet~\cite{lutnet1}) and LUT-based traditional networks (LogicNets~\cite{logicnets}, NullaNet~\cite{nullanet}). 

Wang \textit{et al.} introduced LUTNet, the first neural network architecture in which the BNNs’ XNOR operations are replaced by learned $K$-input Boolean operations~\cite{lutnet1}. By using the full flexibility of the LUTs, LUTNet reaches high logic density, allowing heavy network pruning. However, the design is bottlenecked by large adder trees and the number of parameters that need to be trained scales exponentially with the number of LUT inputs. In contrast, LogicNets and NullaNet are two architectures that encapsulate all the operations that happen between the quantized inputs and the quantized outputs of a traditional linear + activation neuron, enumerating the function values, and hence generating a netlist of logical-LUTs (L-LUTs) which get mapped to physical-LUTs (P-LUTs) by the logic synthesis tools.

We expand on the idea of converting a neural network into a netlist of LUTs by exploiting the LUTs' ability to implement a wide range of functions. A neural network consisting of linear layers and rectified linear units taken as a whole computes a continuous piecewise linear function. Rather than training these traditional functions, we suggest training continuous piecewise polynomial functions. This approach offers the advantage of providing a high degree of flexibility in fitting training data while maintaining unchanged truth table sizes. We are able to show that significantly shallower, and hence lower-latency networks are possible as a result. We perform the polynomial transform inside each neuron by augmenting the feature space with all of its monomials of degree at most some tunable constant, $D$, providing a tunable trade-off between generality, potential for overfitting, and the number of training parameters.

Our aim is to enable applications that require ultra-low latency real-time processing and highly lightweight on-chip implementations. Examples include the classification of particle collision events~\cite{duarte}, the detection of malicious network traffic~\cite{murovic}, and image classification for real-time systems, hence in common with other work in this field, we focus on networks that can be implemented entirely on a single FPGA.
\begin{figure*}[!b]
     \centering
     \hspace{-9mm}
     \begin{subfigure}[b]{0.3\textwidth}
         \captionsetup{width=.9\textwidth}
         \includegraphics[width=\textwidth]{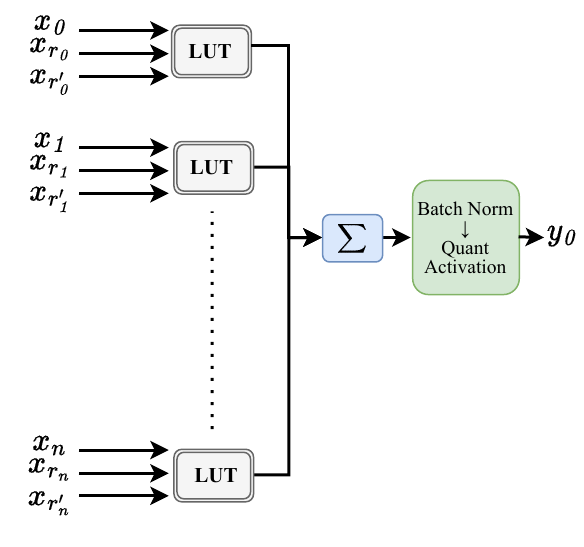}
         \caption{LUTNet - $K$-LUTs, here $3$-LUTs, replace the XNORs present in a BNN. $x$ would be the input vector to the XNORs. $x_r$ and $x_{r'}$ are random selections (with replacement) from $x$.}
         \label{fig:lutnet}
     \end{subfigure}
     \hspace{4mm}
     \begin{subfigure}[b]{0.265\textwidth}
         \captionsetup{width=.8\textwidth}
         \includegraphics[width=\textwidth]{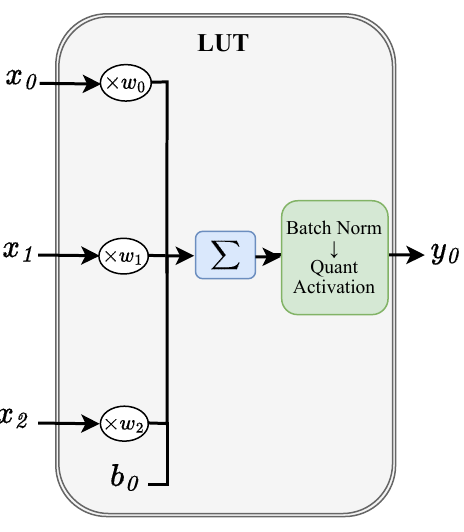}
         \caption{LogicNets - L-LUTs are used to absorb all the operations between the quantized inputs and quantized output.\\}
         \label{fig:logicnets}
     \end{subfigure}
     \hspace{3.5mm}
     \begin{subfigure}[b]{0.37\textwidth}
         \captionsetup{width=.75\textwidth}
         \includegraphics[width=\textwidth]{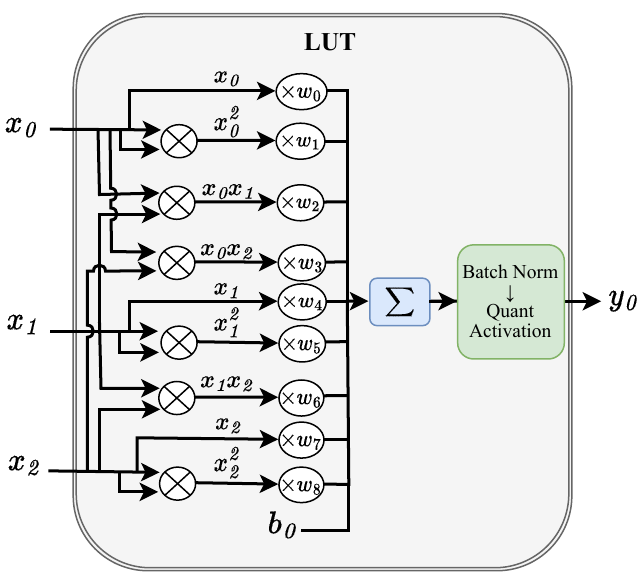}
         \caption{PolyLUT - L-LUTs are used in the same context as LogicNets, however, inside the neuron a multivariate polynomial of degree $D$, here $D=2$, replaces the linear transformation.}
         \label{fig:polylut}
     \end{subfigure}
     \hspace{-8mm}
        \caption{Structural use of LUTs for a single output channel in prior works (a,b) and our approach (c).}
        \label{fig:three graphs}

\end{figure*}

This paper makes the following novel contributions:
\begin{itemize}
    \item{We present an open source framework called PolyLUT, the first quantized neural network training methodology that maps a neuron to a logical-LUT while using multivariate polynomial function learning to exploit the flexibility of the FPGA soft logic.\footnote{\url{https://github.com/MartaAndronic/PolyLUT}}}
    \item{We expose the degree of the monomials as a training hyperparameter that can be adjusted to increase the flexibility of the functions. We show empirically that higher degrees increase the ability of the model to fit the training data which allows network depth reductions.}
    \item{For the chosen datasets, we demonstrate the effectiveness of PolyLUT at deployment, achieving shallower networks for the same accuracy, resulting in network inference times of $\sim10\text{ns}$ and Pareto optimal resource utilization. For example, on the jet substructure classification task, we demonstrate a 2-layer piecewise polynomial network achieving the same accuracy as a 4-layer piecewise linear network while offering a significant $2.69\times$ area compression combined with a $2.29\times$ reduction in latency.
    }
\end{itemize}

\section{Background}
The migration of deep learning models from the cloud to the edge presents challenges attributed to limited computational resources~\cite{edge}. To address these challenges, current research is primarily focused on employing various techniques such as compression, algorithm-hardware co-design, and hardware accelerators. Table~\ref{table:prior} summarizes techniques used in prior low-latency FPGA implementations.

\subsection{Low latency FPGA accelerators}
FINN is a framework originally designed for building BNN accelerators that reached the fastest classification rates on MNIST at the time of first publication~\cite{finn}. Ngadiuba \textit{et al.}~\cite{hls4ml} also implement fully unrolled binary and ternary neural networks in the \texttt{hls4ml} package, which maps neural networks to FPGA circuits to target reduced computation time and resource utilization. In contrast, Duarte \textit{et al.}~\cite{duarte} use \texttt{hls4ml} but adopt higher network precision to generate accurate and fast fully-unrolled and rolled designs at the cost of high digital signal processing (DSP) block usage. Fahim \textit{et al.} build upon these two \texttt{hls4ml} works and also leverage the insights from prior approaches using boosted decision trees~\cite{summers} and quantization-aware training~\cite{coelho}. Additionally, Fahim \textit{et al.} introduce new low power techniques, such as quantization-aware pruning to enhance performance and efficiency. Another real-time implementation of BNNs is presented by Murovic \textit{et al.}~\cite{murovic} who achieve high performance by exploiting full parallelization.

\begin{table}[tbp]
\caption{Key features of prior low-latency accelerators.}
\begin{center}
\renewcommand{\arraystretch}{1.2} 
\begin{tabular}{|c|l|l|l|}
\Xhline{2\arrayrulewidth}
\multicolumn{1}{|c|}{\multirow{2}{*}{\textbf{Tool}}}&\multicolumn{1}{|c|}{\multirow{2}{*}{\textbf{Precision}}}&\multicolumn{1}{|c|}{\multirow{2}{*}{\textbf{Pruning}}}&\multicolumn{1}{|c|}{\textbf{Implementation}}\\\multicolumn{1}{|c|}{} &\multicolumn{1}{|c|}{}&\multicolumn{1}{|c|}{}&\multicolumn{1}{|c|}{\textbf{type}}\\
\Xhline{2\arrayrulewidth}
FINN&\multirow{2}{*}{Binary}&\multirow{2}{*}{None}&\multirow{2}{*}{XNOR-based}\\
~\cite{finn}&&&\\
\hline
Ngadiuba&Binary/ &\multirow{2}{*}{None}&\multirow{2}{*}{XNOR-based}\\
\textit{et al.}~\cite{hls4ml}&Ternary&&\\
\hline
Duarte&\multirow{2}{*}{Fixed point}&\multirow{2}{*}{Iterative pruning}&\multirow{2}{*}{DSP-based}\\
\textit{et al.}~\cite{duarte}&&&\\
\hline
Fahim&\multirow{2}{*}{Fixed point}&Quantization-aware&\multirow{2}{*}{DSP-based}\\
\textit{et al.}~\cite{fahim}&&iterative pruning&\\
\hline
Murovic&\multirow{2}{*}{Binary}&\multirow{2}{*}{None}&\multirow{2}{*}{XNOR-based}\\
\textit{et al.}~\cite{murovic}&&&\\
\hline
LUTNet&Residual&Pruning on&\multirow{2}{*}{LUT-based}\\
~\cite{lutnet1}&binarization&full-precision model&\\
\hline
LogicNets&\multirow{2}{*}{Low-bit}&\textit{A priori}&\multirow{2}{*}{LUT-based}\\
~\cite{logicnets}&&fixed sparsity&\\
\hline
\end{tabular}
\renewcommand{\arraystretch}{1.2}
\label{table:prior}
\end{center}
\end{table}
\vspace{0.2cm}
\subsection{BNN-specific micro-architectures using FPGA LUTs}
\subsubsection{Differentiable LUTs} LUTNet replaced the dot product between the feature vector and weight vector with a sum of learned Boolean functions that can each be computed using a single $K$-LUT. The structure of LUTNet can be visualized in Figure~\ref{fig:lutnet}, where $K=3$ and $x_r$ and $x_{r'}$ are vectors generated by randomly selecting (with replacement) from the input vector $x$. Therefore, in LUTNet, each input can participate multiple times in the weighted summation and redundant operations can be reduced with the help of network pruning, while maintaining accuracy~\cite{lutnet1}. 

Compared to PolyLUT, LUTNet contains exposed datapaths within neurons, exhibits exponential scaling of training parameters with LUT input size, and supports a single level of quantization.
\vspace{0.2cm}
\subsubsection{LUT-based traditional networks} NullaNet~\cite{nullanet} proposes viewing layers as multi-input multi-output Boolean functions that can be optimized through Boolean logic minimization. To address the exponential growth of the size of the truth tables, output values are determined only for a strict subset of input combinations with logic synthesis treating the others as don't-care conditions. LogicNets~\cite{logicnets} is designed for very low latency processing. LogicNets uses high sparsity to reduce the large fan-in in order to combat the disadvantage of NullaNet’s lossy truth table sampling method. Our work utilizes the same \textit{a priori} fixed sparsity technique as LogicNets, based on expander graph theory~\cite{expander}. Figure~\ref{fig:logicnets} shows the structure of a LogicNets neuron. Due to sparsity, the input vector $x$ is of size $F \ll N$, with $F=3$ in the figure. After training, the function that gets absorbed inside the L-LUT can be expressed as in (\ref{logicnets_lut}), where $\phi$ is the quantized activation function, $w$ is the real weight vector,  and $b$ is the real bias term.
\begin{equation}
\label{logicnets_lut}
y_0=\phi\left[\sum_{i=0}^{F}w_ix_i + b\right]
\end{equation}

\section{Methodology}
\label{section:methodology}
The novelty of our work comes through the expansion of the feature vector $x$ at each neuron with all the monomials up to a parametric degree. Figure~\ref{fig:polylut} illustrates PolyLUT's expansion for $D=2$. When $D=1$, PolyLUT's behavior becomes linear and it is equivalent to LogicNets. It is thus a strict generalization of LogicNets.

\subsection{Theoretical approach}

Provided with a sufficient number of neurons, a deep neural network is capable of modeling any function within a given error margin. However, there is theoretical evidence~\cite{multi} that multiplicative interactions are a way of integrating contextual information in a neural network. It is proven that the hypothesis space of a traditional multi-layer perceptron with rectified linear unit activation functions is strictly included in the hypothesis space of an equivalent network with each linear layer replaced by a multiplicative layer~\cite{multi}. Therefore, multiplicative interactions enlarge the hypotheses space allowing for improved representational power, improved modeling of interactions and robustness.

In a direct implementation, integrating multiplicative interactions in the neural network would increase the number of multipliers used, {\em however, in a LUT-based model, this additional complexity is entirely absorbed inside the logical-LUT, and no additional multiplication hardware is required}. This is the key to our approach, leading to a significant boost in model performance with very little overhead. We are able to recycle the increase in function complexity within each layer to reduce the number of layers required to achieve a given accuracy.

Our fully-trainable layers can be described in the following way. Consider a vector $x$ of inputs to a standard neural network layer. We may first form all multiplicative combinations of these inputs up to a parametric degree $D$. For example, if a given input vector is two-dimensional and $D=3$, then model construction arises from the following transformation: $[x_0,x_1]\mapsto[1,x_0,x_1,x_0^2,x_0x_1,x_1^2,x_0^3,x_0^2x_1,x_0x_1^2,x_1^3]$. The size of the feature vectors will see an increase from $F$ to $\frac{(F+D)!}{F!D!}$, where $D$ can be seen as a hyperparameter allowing a smooth variation from a number of terms linear in $F$ when $D=1$, equivalent to LogicNets, to a number of terms exponential in $F$ when $D=F$, akin to LUTNet, as seen in Table~\ref{table:params}. After this transformation, we apply a standard linear layer + activation to the resulting expanded feature vector. This can therefore be seen as replacing (\ref{logicnets_lut}) with (\ref{polylut_lut}), where $M$ is equal to the number of monomials $m(x)$ of degree at most $D$ in $F$ variables.
\begin{equation}
\label{polylut_lut}
y_0=\phi\left[\sum_{i=0}^{M}w_im_i(x)\right]\text{, where M} = \left(F+D \atop D \right)
\end{equation}

\begin{table}[t]
\caption{Number of parameters for a $\beta F$-input LUT. $\beta$ is the input bitwidth and $F$ the neuron fan-in. PolyLUT's number of parameters lies between Logicnets' and LUTNet's number of parameters. In contrast to LUTNet, LogicNets' and PolyLUT's number of parameters is independent of the input bitwidth.}
\begin{center}
\begin{tabular}{|c|c|c|}
\hline
&\multirow{2}{*}{\textbf{Number of parameters}}&\multirow{2}{*}{\textbf{Scaling type}}\\&&\\
\hline
\multirow{2}{*}{\textbf{LUTNet}}&\multirow{2}{*}{$2^{\beta F}$}&\multirow{2}{*}{Exponential}\\&&\\
\hline
\multirow{2}{*}{\textbf{PolyLUT}}&\multirow{2}{*}{$\left(F+D \atop D \right)$}&Polynomial for \\&&fixed D\\
\hline
\multirow{2}{*}{\textbf{LogicNets}}&\multirow{2}{*}{$F+1$}&\multirow{2}{*}{Linear}\\&&\\
\hline

\end{tabular}
\label{table:params}
\end{center}
\end{table}

A common activation function in deep learning is the rectified linear unit (ReLU). From a functional perspective, each ReLU partitions the input space into two linear sub-regions. As the network depth increases, more piecewise linear surfaces are generated.

In the polynomial setting, the input space is transformed into a surface described by a multivariate polynomial function. This polynomial transformation offers significantly greater flexibility and expressive power compared to a linear transformation. As a result, a network that incorporates polynomial transformations can accurately capture complex relationships in the data using fewer ``folds" or neurons.

\begin{figure}[!t]
\centerline{\includegraphics[width=90mm]{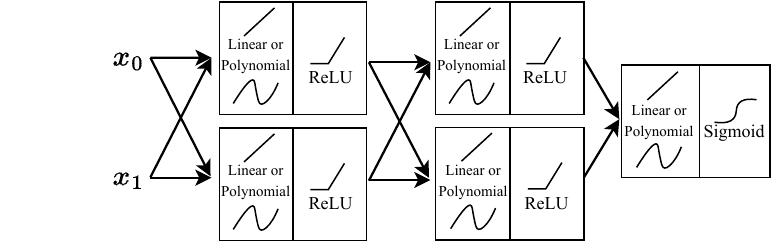}}
\caption{Visual representation of a 3-layer network.}
\label{fig:net}
\end{figure}
\vspace{0.2cm}

\begin{figure}[!t]
     \centering
     \begin{subfigure}[b]{0.49\textwidth}
         \includegraphics[width=\textwidth]{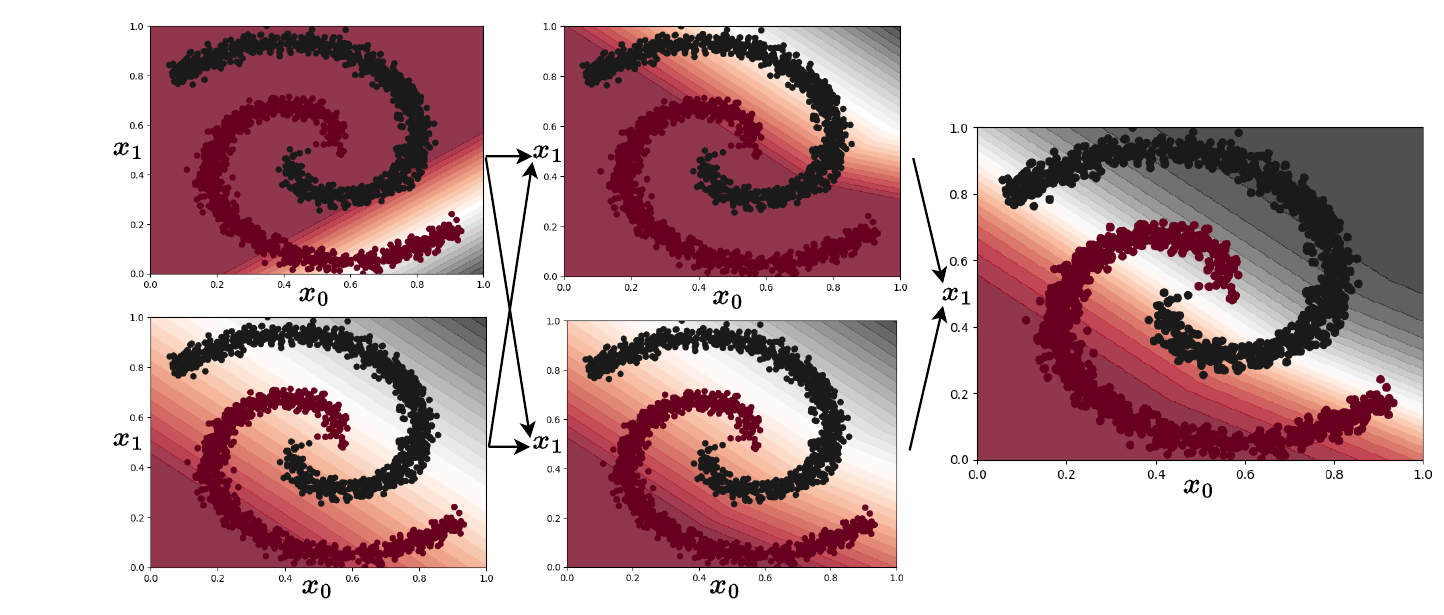}
         \caption{Continuous piecewise linear function.}
         \label{fig:linear}
     \end{subfigure}
     \hspace{2mm}
     \begin{subfigure}[b]{0.49\textwidth}
         \includegraphics[width=\textwidth]{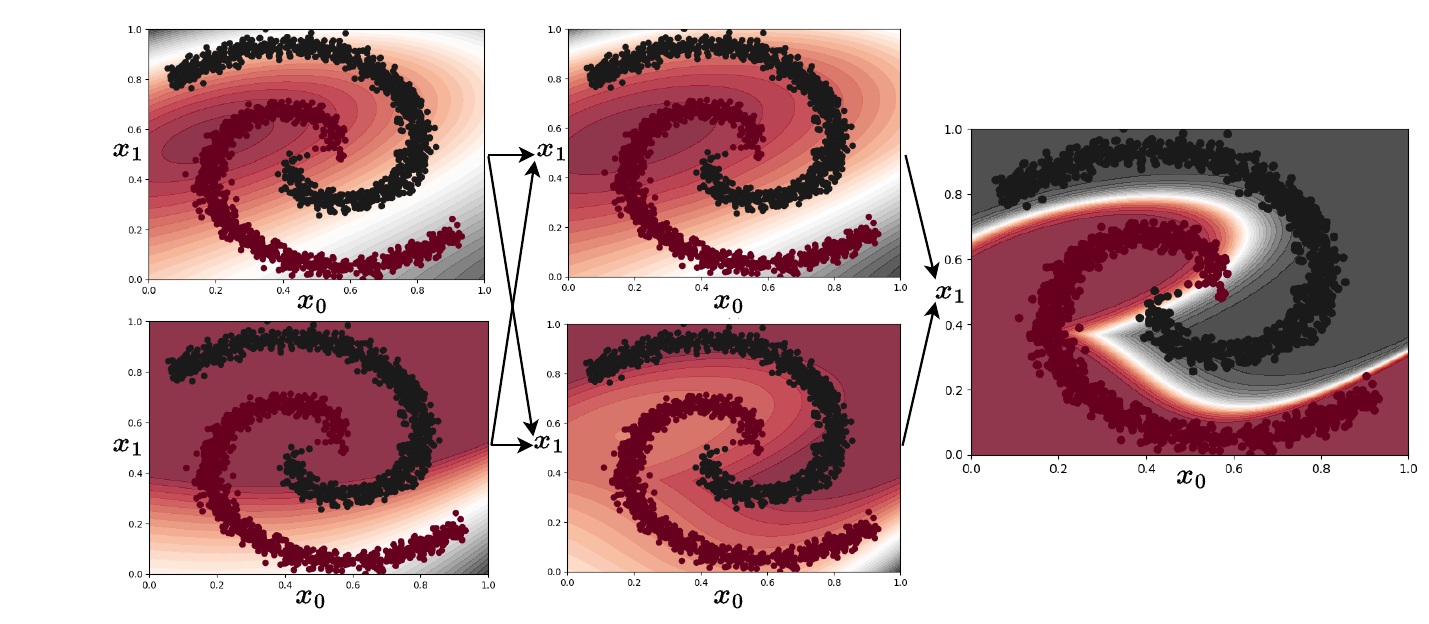}
         \caption{Continuous piecewise polynomial function.}
         \label{fig:poly}
     \end{subfigure}
        \caption{Input transformations visualized as contour graphs at the output of each neuron. The black and red dots represent the training data points.}
        \label{fig:two graphs}
\end{figure}
To visually illustrate this concept, we present a two-dimensional toy network in Figure~\ref{fig:net}, which we have trained in two different settings. The first setting incorporated linear functions and the second setting polynomial functions of degree equal to three. The resulting transformations can be observed in Figure~\ref{fig:two graphs}. We have used a fictional dataset of two spirals, each representing a class (black and red). 

In Figure~\ref{fig:linear}, the formation of the piecewise linear surface can be visualized at the output of each neuron, as a contour graph. As expected, the first layer of neurons tries to separate the data using a single line. Following, the second layer brings in another linear transformation that results in two separate lines on the contour graph. Consequently, the final classification contains four distinct linear shapes and the darker shades express an increased level of confidence. The importance of the ReLU function can also be visualized, which prevents the network from collapsing into one single linear function. Looking at the final piecewise linear classification it is possible to observe that four linear functions are insufficient to separate the data faithfully due to the high degree of overlap in the training data. In order to increase the accuracy of the classification there would be a need to increase the width or depth of the network. 

In Figure~\ref{fig:poly}, the formation of the piecewise polynomial surface is shown. It is evident from the graphs that the piecewise polynomial surface can adopt more intricate shapes, enabling it to distinguish more complex data patterns. The network is able to fit the training data with significantly better accuracy, being able to separate the intertwined spirals. In contrast to the linear case, which is implemented in LogicNets, the polynomial model introduces enhanced data fitting capabilities, facilitating better representation of the underlying data patterns.

A high-degree multivariate polynomial model introduces more degrees of freedom, which can lead to overfitting. To mitigate these issues, two regularization techniques can be employed: reducing the number of layers or decreasing the degree of the polynomial. The former leads to better generalization through the reduction of the number of polynomial regions. The latter increases the generalization while maintaining the same number of layers. By controlling the model's complexity through layer reduction, significant resources can be saved. Moreover, this approach leads to a reduction in the number of clock cycles required to classify input data, thereby improving computational efficiency.

LUTNet and PolyLUT differ in several key aspects, one of which is the number of operations absorbed inside the LUTs. With PolyLUT, not only is the weighted feature vector absorbed, but also the summation, activation, and quantization functions, resulting in no exposed datapath within a neuron. Another notable difference is the choice of real functions used for network training. LUTNet trains LUT functions using the fixed degree Lagrange interpolating polynomial, whereas PolyLUT trains the coefficients of each monomial. Consequently, PolyLUT enables variations in function complexity through degree adjustment during training and allows the reduction of the trainable parameters, as per Table~\ref{table:params}. Moreover, the exponential increase in the number of parameters in LUTNet is dependent on bitwidth ($\beta$). The final key difference is that LUTNet operates solely in the binary space, while PolyLUT supports any level of quantization.

\subsection{Hardware approach}
A lookup table is capable of evaluating any function with quantized inputs and outputs. For a fixed number of LUT inputs, the time complexity of evaluation is independent of the complexity of the function stored in the LUT, making it ideal for implementing computationally intensive functions. By mapping complex computations to lookup tables, their full potential can be leveraged.

The primary component of the soft logic in an FPGA is the $K$-LUT, which can implement any Boolean function with $K$ inputs and one output. Utilizing truth tables to represent the computation of an artificial neuron proves highly efficient as it combines multiple operations such as multiplication, summation, activation, quantization, and normalization into a single operation that can be executed within a single clock cycle or less.

It is important to note that the size of a lookup table grows exponentially with the number of inputs. To address this, we follow LogicNets by introducing extreme sparsity, limiting the number of inputs a neuron can accept to a fan-in of $F$ and we apply heavy quantization, constraining the bit-width of the inputs to $\beta<4 \text{ bits}$. As a result, the size of the truth tables becomes $\mathcal{O}(2^{\beta\cdot F})$. The complexity of the truth tables is therefore established through a trade-off between the precision of data representation and the sparsity of the network.

\subsection{Building a neural network of LUTs}
\begin{figure}[t]
\centerline{\includegraphics[width=60mm]{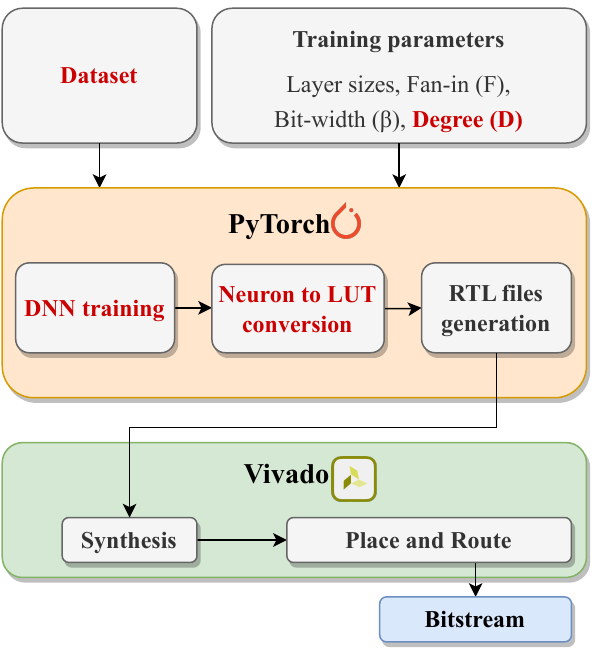}}
\caption{High-level view of PolyLUT's toolflow. We built upon the open-source LogicNets toolflow. We show in red the elements that were modified.}
\label{fig:flow}
\end{figure}

PolyLUT is a deep neural network (DNN) co-design methodology that builds upon LogicNets' toolflow~\cite{logicnets} which realizes the training of quantized DNNs, conversion to truth-tables, and generation of hardware netlists. The training implementation was changed to accommodate PolyLUT. The toolflow's complete structure can be visualized in Figure~\ref{fig:flow}.

\subsubsection{DNN Training}
For the DNN training process, we utilize a PyTorch framework. Upon training, certain specifications are required from the user, including the dataset, model layer sizes, fan-in, bit-width, degree, and other machine learning-specific parameters such as learning rate and weight decay.

The Pytorch framework includes a parameter, \texttt{degree}, to control the complexity of the monomials. This parameter allows for adjusting the complexity of the model and can also be set to one to enable linear learning. We train our model using Decoupled Weight Decay Regularization~\cite{reg} and Stochastic Gradient Descent with Warm Restarts~\cite{sgd}. The inputs and outputs of each layer are batch normalized and quantized using the Brevitas~\cite{brevitas} quantized activation functions, which employ learned scaling factors.

\subsubsection{Neuron to L-LUT conversion}
After the model has been trained, the next step involves calculating the truth tables for each neuron. This process entails generating all possible input combinations based on their specified bit-width and fan-in, and then tabulating the neuron function evaluated on these combinations to obtain the corresponding outputs. As a result, each neuron requires the evaluation of $2^{\beta\cdot F}$ input combinations to construct its truth table. 

\subsubsection{RTL files generation}
Each L-LUT is converted to Verilog RTL, by writing out the truth tables as read-only memories (ROMs) that are converted to P-LUTs by the synthesis tool. Additionally, each ROM is coded with registers at the output and the input connections are specified using the activation masks.

\section{Experimental Results}
\subsection{Datasets}
To evaluate the efficacy of our method, we have analyzed three datasets that are commonly used for the evaluation of ultra-low latency and size-critical architectures. The first one is the UNSW-NB15 dataset~\cite{unsw}, the second one is the MNIST dataset~\cite{mnist} and the final one is the jet tagging dataset as presented in~\cite{duarte}, which has also been used for a case study.
\vspace{0.2cm}
\subsubsection{Network intrusion detection}
Ultra-low latency DNNs find application in cybersecurity, particularly in network intrusion detection systems (NIDS). These systems are responsible for monitoring network traffic and identifying and protecting against malicious packets. Given that fiber-optic internet can achieve speeds of up to $940$ Mbps, it is essential to have NIDS solutions that can handle such high throughput. However, privacy is also a significant concern in this context, making on-chip FPGA design an ideal choice for implementation. To showcase the effectiveness of our work in this domain, we have trained our models on the UNSW-NB15 dataset, which has been randomly partitioned as described in~\cite{murovic}. This dataset is designed for the purpose of labeling network packets as either safe ($0$) or malicious ($1$), based on $49$ input features.
\vspace{0.2cm}
\subsubsection{Handwritten digit recognition}
The advancements in autonomous driving, augmented reality, and edge technologies have created a demand for real-time image classification. The ability to perform low-latency inference opens up a wide range of applications in this rapidly evolving field. To assess our work, we have chosen the MNIST dataset, which aligns with the capabilities of our networks. The MNIST dataset consists of handwritten digits represented as $28\times28$ pixel images. The input pixels are flattened, resulting in inputs of size $784$, while the $10$ output classes correspond to each digit.
\vspace{0.2cm}
\subsubsection{Jet substructure classification}
Real-time inference and efficient resource utilization have played a crucial role in driving physics advancements at the CERN Large Hadron Collider (LHC). Given the high collision frequencies at the LHC, the initial stage of data processing demands exceptional throughput to effectively refine large volumes of sensor data. Prior works~\cite{duarte,fahim,hls4ml,summers,coelho} employed neural networks to deploy inference models on FPGAs for the task of jet substructure analysis. This particular task involves processing $16$ substructure properties to classify $5$ types of jets. We showcase our work on this task as well, showing that PolyLUT could be a solution for applications in high-energy physics.
\subsection{Training and implementation setup}
All the models are trained for $1000$ epochs for the smaller datasets, Jet Substructure and UNSW-NB15, and for $500$ epochs for the MNIST dataset. We use a mini-batch of $1024$ and $256$, respectively. Additionally, the models from this paper were chosen to fit on a single FPGA device.
To compile the Verilog files generated by the PyTorch script we use Vivado 2020.1 and, to keep consistency with LogicNets, we target the $\texttt{xcvu9p-flgb2104-2-i}$ FPGA part. Also for consistency, the projects are compiled using the Vivado \texttt{Flow\_PerfOptimized\_high} settings, and the projects are configured to perform synthesis in the \texttt{Out-of-Context} (OOC) mode. We target a clock period of $1.6$ ns for the smaller networks and $2$ ns for the largest ones. The accuracy of the netlists is verified using the LogicNets' pipeline since the hardware architecture remains unchanged.

\subsection{Case study}
As a case study, we selected the jet substructure tagging task to investigate the impact of the degree on training loss, test accuracy, latency, and area. Our objective is to demonstrate that incorporating increased function complexity through polynomial expansion improves training performance and enables ultra-low latency FPGA implementations. By focusing on the jet substructure tagging task, we are able to gain insights into the broader implications and benefits of our methodology.

As our baseline, we adopted the JSC-M architecture from LogicNets, as outlined in Table~\ref{table:networks}. The JSC-M architecture comprises five layers with parameters $\beta=3$ and $F=4$.

For our experimentation, we trained the JSC-M network using polynomial degrees ranging from 1 to 6. Additionally, we systematically eliminated one hidden layer at a time, generating a set of sub-networks that permitted the evaluation of the same range of polynomial degrees. This approach allowed us to analyze the effects of layer elimination on the performance metrics mentioned above. Since each layer in the network can be processed within a single clock cycle, we specifically considered networks that exhibit varying inference speeds, ranging from $2$ to $5$ clock cycles. This ensured that we examined the impact of different network configurations on both accuracy and efficiency, taking into account the trade-off between inference speed and test accuracy.
\begin{table*}[htbp]
\caption{Model architectures used as benchmarks for evaluated datasets.}
\begin{center}
\renewcommand{\arraystretch}{1.2} 
\begin{tabular}{cccccc}
\hline
\textbf{Dataset}&\textbf{Model Name}&\textbf{Nodes per Layer}&$\boldsymbol{\beta}$&\textbf{$F$}&\textbf{Exceptions}\\
\hline
Jet substructure & JSC-M & 64, 32, 32, 32, 5 & 3 & 4& \\
\hline
Jet substructure & JSC-M Lite & 64, 32, 5 & 3 & 4& \\
\hline
Jet substructure & JSC-XL & 128, 64, 64, 64, 5 & 5 & 3& $\beta_0=7$, $F_0=2$\\
\hline
UNSW-NB15 & NID Lite & 686, 147, 98, 49, 1 & 2 & 7&$\beta_0=1$\\
\hline
MNIST & HDR & 256, 100, 100, 100, 100, 10 & 2 & 6& \\
\hline

\end{tabular}
\renewcommand{\arraystretch}{1.2} 
\label{table:networks}
\end{center}
\end{table*}

\vspace{0.2cm}
\subsubsection{The impact of degree on training loss and test accuracy}
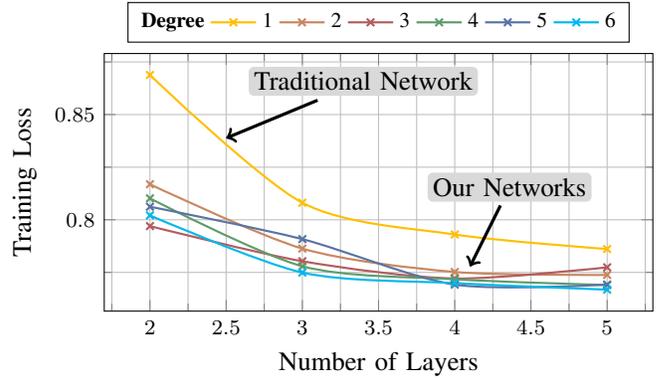
\begin{figure}[tbp]
	\centering
        \hspace{-1.1cm}
        \resizebox{1.1\columnwidth}{!}{\definecolor{myblue}{rgb}{0.35, 0.45, 0.64}
\definecolor{myorange}{rgb}{0.79, 0.53, 0.38}
\definecolor{myred}{rgb}{0.7, 0.36, 0.37}
\definecolor{mygreen}{rgb}{0.37, 0.61, 0.42}
\definecolor{mymagenta}{rgb}{0.52, 0.47, 0.66}
\definecolor{myyellow}{rgb}{1.0, 0.75, 0.0}
\definecolor{mycyan}{rgb}{0.0, 0.72, 0.92}
\definecolor{codegreen}{rgb}{0,0.6,0}

\begin{tikzpicture}
\pgfplotsset{compat=1.5}
\begin{axis}[
  width=1\columnwidth,
  height=50mm,
  grid=both,
  legend columns=-1,
  minor x tick num=1,
  minor y tick num=1,
  xlabel=Number of Layers,
  ylabel=Training Loss,
  tick label style={font=\footnotesize},  
  legend style={at={(0.5,1.2)},anchor=north, font=\footnotesize},
  every node/.style={
    font=\sffamily\scriptsize
    },  
    circtext/.style={draw,circle,minimum size=8pt,inner sep=2pt},
    dot/.style={draw,circle,fill=black,minimum size=0.6mm,inner sep=0pt},
]

\addlegendimage{empty legend}

\addplot [myyellow, smooth, thick, mark color=myblue,
mark=x]coordinates {
    (02, 0.8688)
    (03, 0.8081)
    (04, 0.793)
    (05, 0.7861)
};

\addplot [myorange,smooth, thick, mark color=myorange,
mark=x] coordinates {
    (02, 0.8169)
    (03, 0.7863)
    (04, 0.7752)
    (05, 0.7738)
};
\addplot [myred,smooth,thick,mark color=myred,
mark=x] 
coordinates {
    (02, 0.797)
    (03, 0.7803)
    (04, 0.7721)
    (05, 0.7774)
};
\addplot [mygreen,smooth,thick,mark color=mygreen,
mark=x] 
coordinates {
    (02, 0.8102)
    (03, 0.7779)
    (04, 0.7717)
    (05, 0.769)
};
\addplot [myblue,smooth,thick,mark color=myyellow,
mark=x] 
coordinates {
    (02, 0.8062)
    (03, 0.7908)
    (04, 0.7692)
    (05, 0.7692)
};
\addplot [mycyan,smooth,thick,mark color=mycyan,
mark=x] 
coordinates {
    (02, 0.8021)
    (03, 0.7749)
    (04, 0.7699)
    (05, 0.7668)
};

\draw[black,very thick,<-] (50,72) -- (110,90) node [fill=gray!30!white,pos=1.5, rounded corners=2pt, inner sep=2pt]
{Traditional Network};

\draw[black,very thick,<-] (210,11) -- (230,40) node [fill=gray!30!white,pos=1.3, rounded corners=2pt, inner sep=2pt]
{Our Networks};

\addlegendentry{\hspace{-.0cm}\textbf{Degree}}\addlegendentry{1}
\addlegendentry{2}
\addlegendentry{3}
\addlegendentry{4}
\addlegendentry{5}
\addlegendentry{6}
\end{axis}
\end{tikzpicture}}
	\caption{
            Training loss variation with the number of layers across six different polynomial degrees.
        }
	\label{fig:g0}
\end{figure}

\begin{figure}[tbp]
	\centering
 \hspace{-1.1cm}
        \resizebox{1.1\columnwidth}{!}{\definecolor{myblue}{rgb}{0.35, 0.45, 0.64}
\definecolor{myorange}{rgb}{0.79, 0.53, 0.38}
\definecolor{myred}{rgb}{0.7, 0.36, 0.37}
\definecolor{mygreen}{rgb}{0.37, 0.61, 0.42}
\definecolor{mymagenta}{rgb}{0.52, 0.47, 0.66}
\definecolor{myyellow}{rgb}{1.0, 0.75, 0.0}
\definecolor{mycyan}{rgb}{0.0, 0.72, 0.92}
\definecolor{codegreen}{rgb}{0,0.6,0}

\begin{tikzpicture}
\pgfplotsset{compat=1.5}
\begin{axis}[
  width=1\columnwidth,
  height=50mm,
  grid=both,
  legend columns=-1,
  minor x tick num=1,
  minor y tick num=1,
  xlabel=Number of Layers,
  ylabel=Test Accuracy (\%),
  tick label style={font=\footnotesize},  
  legend style={at={(0.5,1.2)},anchor=north, font=\footnotesize},
]
\addlegendimage{empty legend}

\addplot [myyellow,smooth, thick, mark color=myblue,
mark=x]coordinates {
    (02, 69.73)
    (03, 71.4)
    (04, 71.78)
    (05, 71.86)
};
\addplot [myorange,smooth, thick, mark color=myorange,
mark=x] coordinates {
    (02, 70.91)
    (03, 72.15)
    (04, 72.39)
    (05, 72.3)
};
\addplot [myred,smooth,thick,mark color=myred,
mark=x] 
coordinates {
    (02, 71.79)
    (03, 72.1)
    (04, 72.43)
    (05, 72.32)
};
\addplot [mygreen,smooth,thick,mark color=mygreen,
mark=x] 
coordinates {
    (02, 71.42)
    (03, 72.27)
    (04, 72.42)
    (05, 72.46)
};
\addplot [myblue,smooth,thick,mark color=myyellow,
mark=x] 
coordinates {
    (02, 71.64)
    (03, 72.01)
    (04, 72.37)
    (05, 72.41)
};
\addplot [mycyan,smooth,thick,mark color=mycyan,
mark=x] 
coordinates {
    (02, 71.36)
    (03, 72.29)
    (04, 72.33)
    (05, 72.46)
};

\addplot[densely dashed, black, thick]
coordinates {(02, 71.86) ((05, 71.86)};

\draw[black,very thick,<-] (25,35) -- (100,35) node [fill=gray!30!white,pos=1.5, rounded corners=2pt, inner sep=2pt]
{Traditional Network};

\draw[black,thick,densely dashed, <-] (140,208) -- (170,100) node[fill=gray!30, text=black, font=\scriptsize, rounded corners=2pt, inner sep=1pt] {$\approx$ same accuracy with $< 0.5\times$ layers};;

\draw[black,very thick,<-] (210,250) -- (240,180) node [fill=gray!30!white,pos=1.3, rounded corners=2pt, inner sep=2pt]
{Our Networks};

\addlegendentry{\hspace{-.0cm}\textbf{Degree}}
\addlegendentry{1}
\addlegendentry{2}
\addlegendentry{3}
\addlegendentry{4}
\addlegendentry{5}
\addlegendentry{6}
\end{axis}
\end{tikzpicture}}
	\caption{
            Test accuracy variation with the number of layers across six different polynomial degrees.
        }
	\label{fig:g1}
\end{figure}
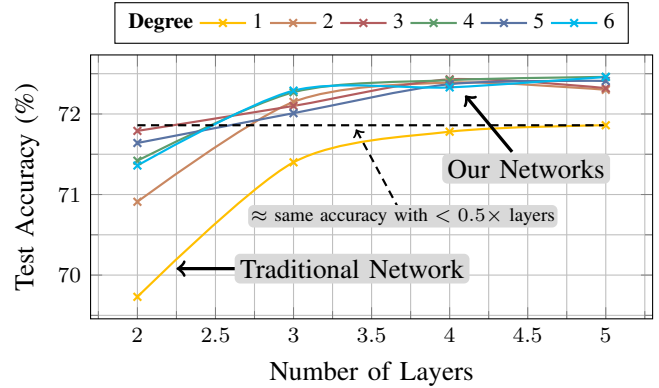
Firstly, we analyze the effect of gradually increasing the degree on both training loss and test accuracy. The results of this analysis can be visualized in Figure~\ref{fig:g0} and Figure~\ref{fig:g1}. 

Observing the trends in training loss, it is evident that the curve representing the points with $D=1$ is positioned above all the higher-degree curves. This indicates that models with $D=1$ are less efficient in minimizing loss compared to the other data points. We observe a consistent improvement as the degree increases from $1$ to $2$. However, as the degree is further increased, the improvements become less consistent, suggesting that very high degrees may bring diminishing returns. This can be explained by the fact that increasing the function complexity allows for better fitting of the data. However, larger degrees tend to be more challenging to train.

Upon analyzing the test accuracy, we find that the same trends as observed in the training loss apply. However, in this case, the separation between the data points with higher degrees appears to be even more unpredictable, suggesting that large degrees can result in overfitting the training data.

The most valuable observation from these graphs is that the performance boost achieved through polynomial function training is equivalent to adding more layers to a linear network. In this study, we observe that training a linear $5$-layer network yields lower accuracy compared to training a $3$-layer polynomial network, and the accuracy is nearly the same as that of a $2$-layer polynomial network.

These findings highlight the effectiveness of multivariate polynomial function training in improving network performance, allowing for comparable or even better accuracy with fewer layers compared to linear networks. Thus, the flexibility introduced by polynomial functions can compensate for training very deep networks, resulting in more efficient models.

\vspace{0.2cm}
\subsubsection{The impact of degree on latency and resource utilization}
In the previous section, it was shown that PolyLUT networks allow reductions in the number of layers for the same level of accuracy. Notably, each layer reduction corresponds to a decrease in the inference speed by one clock cycle. 

Figure~\ref{fig:g3} depicts the latency and test error rates of all data points from the previous section. It is evident that the Pareto frontier of the polynomial design points demonstrates greater efficiency in the trade-off between latency and test error rate, surpassing the performance of the Pareto frontier of the points with $D=1$. For the same test error rate, we record reductions in latency ranging from $1.76\times$ to $2.29\times$.

If we were to calculate the theoretical cost using the LUTCost function used in~\cite{logicnets}, it would yield the same result regardless of the polynomial degree. However, in practice, some Boolean LUTs with $N$ inputs will naturally synthesize to smaller circuits than others, when implemented within a network of the native FPGA $K$-LUTs for $N > K$. This can be seen in Figure~\ref{fig:g3} which illustrates the LUT count of all data points from the previous section alongside their respective test error rates. 

\begin{figure}[tbp]
	\centering
 \hspace{-1.1cm}
        \resizebox{1.1\columnwidth}{!}{\definecolor{myblue}{rgb}{0.35, 0.45, 0.64}
\definecolor{myorange}{rgb}{0.79, 0.53, 0.38}
\definecolor{myred}{rgb}{0.7, 0.36, 0.37}
\definecolor{mygreen}{rgb}{0.37, 0.61, 0.42}
\definecolor{mymagenta}{rgb}{0.52, 0.47, 0.66}
\definecolor{myyellow}{rgb}{1.0, 0.75, 0.0}
\definecolor{mycyan}{rgb}{0.0, 0.72, 0.92}
\definecolor{codegreen}{rgb}{0,0.6,0}

\begin{tikzpicture}
\pgfplotsset{compat=1.5}
\begin{axis}[scatter/classes={
  a={mark=*,myyellow, scale=0.8},
  b={mark=*,myorange, scale=0.8},
  c={mark=*,myred, scale=0.8},
  d={mark=*,mygreen, scale=0.8},
  e={mark=*,myblue, scale=0.8},
  f={mark=*,mycyan, scale=0.8}},
  width=1\columnwidth,
  height=50mm,
  grid=both,
  set layers, 
  mark layer=axis tick labels,
  legend columns=-1,
  minor x tick num=1,
  minor y tick num=1,
  xlabel=Number of LUTs,
  ylabel=Test Error Rate (\%),
  tick label style={font=\footnotesize},  
  legend style={at={(0.5,1.2)},anchor=north, font=\footnotesize},
]
\addlegendimage{empty legend}

\addplot [scatter, only marks, scatter src=explicit symbolic]coordinates {
    (2385, 30.27) [a]
    (6261, 28.6) [a]
    (10394, 28.22) [a]
    (14044, 28.14) [a]
};
\addplot [scatter, only marks, scatter src=explicit symbolic] coordinates {
    (4124, 29.09) [b]
    (10768, 27.85) [b]
    (17502, 27.61) [b]
    (24465, 27.7) [b]
};
\addplot [scatter, only marks, scatter src=explicit symbolic] 
coordinates {
    (3870, 28.21) [c]
    (11572, 27.9) [c]
    (17923, 27.57) [c]
    (25822, 27.68) [c]
};
\addplot [scatter, only marks, scatter src=explicit symbolic] 
coordinates {
    (4476, 28.58) [d]
    (10929, 27.73) [d]
    (20894, 27.58) [d]
    (26201, 27.54) [d]
};
\addplot [scatter, only marks, scatter src=explicit symbolic] 
coordinates {
    (3965, 28.36) [e]
    (10764, 27.99) [e]
    (20449, 27.63) [e]
    (26955, 27.59) [e]
};
\addplot [scatter, only marks, scatter src=explicit symbolic] 
coordinates {
    (4442, 28.64) [f]
    (12436, 27.71) [f]
    (19221, 27.67) [f]
    (26066, 27.54) [f]
};

\addplot [red,thick] 
coordinates {
    (2385, 30.27)
    (3830, 30.27)
    (3830, 28.19)
    (10768, 28.19)
    (10768, 27.85)
    (10929, 27.85)
    (10929, 27.73)
    (12436, 27.73)
    (12436, 27.71)
    (17502, 27.71)
    (17502, 27.61)
    (17923, 27.61)
    (17923, 27.57)
    (26066, 27.57)
    (26066, 27.54)
};
\label{pgfplots:label1}
\addplot [black,thick] 
coordinates {
    (2385, 30.27)
    (6261, 30.27)
    (6261, 28.6)
    (10394, 28.6)
    (10394, 28.22)
    (14044, 28.22)
    (14044, 28.14)
};
\label{pgfplots:label2}

\addplot [black, densely dashed, smooth,->]
coordinates {
    (3830, 28.6)
    (5045, 28.75)
    (6261, 28.6)
}
node[
  text=black,
  font=\tiny,
  above,
  yshift=-0.1cm] at (1215,28.6) [pos=0.5,font=\tiny]{$\mathbf{1.6\times}$};;

\addplot [black, densely dashed, smooth,->]
coordinates {
    (3830, 28.22)
    (7112, 28.35)
    (10394, 28.22)
}
node[
  text=black,
  font=\tiny,
  above,
  yshift=-0.1cm] at (1215,28.6) [pos=0.5,font=\tiny]{$\mathbf{2.69\times}$};;

\addplot [black, densely dashed,->]
coordinates {
    (10768, 28.14)
    (12406, 28.11)
    (14044, 28.14)
}
node[
  text=black,
  font=\tiny,
  below,
  yshift=+0.1cm] at (1215,28.6) [pos=0.5,font=\tiny]{$\mathbf{1.3\times}$};;
 
\draw[black,very thick,<-] (42,230) -- (120,230) node [fill=gray!30!white,pos=1.5, rounded corners=2pt, inner sep=2pt]
{Baseline Pareto Frontier};

\draw[black,very thick,<-] (140,25) -- (160,110) node [fill=gray!30!white,pos=1.3, rounded corners=2pt, inner sep=2pt]
{Our Pareto Frontier};

\addlegendentry{\hspace{-.0cm}\textbf{Degree}}
\addlegendentry{1}
\addlegendentry{2}
\addlegendentry{3}
\addlegendentry{4}
\addlegendentry{5}
\addlegendentry{6}
\end{axis}
\end{tikzpicture}}
	\caption{
            Analysis of the trade-off between test error rate and resource utilization for four models with varying numbers of layers across six different polynomial degrees. Our improved Pareto frontier is depicted by (\ref{pgfplots:label1}).
        }
	\label{fig:g2}
\end{figure}
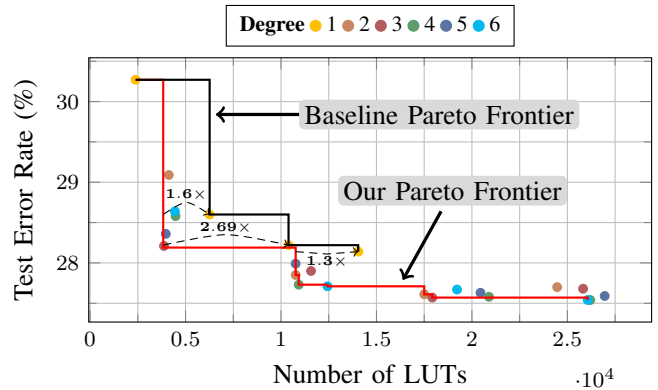

As the complexity increases with higher polynomial degrees slightly more native $K$-LUTs are required for the same network topology. This is a direct consequence of lookup tables now being utilized to express more meaningful functions. However, the increase in LUT utilization due to increasing the degree is well offset by the reduction in the number of layers, overall leading to savings in the number of LUTs. Consequently, as shown in Figure~\ref{fig:g2}, the designs yielded through polynomial training reside on a more efficient Pareto frontier compared to the one formed by the linear networks.

Comparing the performance of a $2$-layer network with $D=3$ (which achieves an accuracy of $71.79\%$) to a $4$-layer network with $D=1$ (which achieves a comparable accuracy of $71.78\%$), we observe a $2.29\times$ improvement in latency and a $2.69\times$ improvement in resource utilization. This demonstrates the advantages of implementing multivariate polynomial functions inside the LUTs in terms of reducing both inference latency and resource utilization.

\begin{figure}[tbp]
	\centering
 \hspace{-1.1cm}
        \resizebox{1.1\columnwidth}{!}{\definecolor{myblue}{rgb}{0.35, 0.45, 0.64}
\definecolor{myorange}{rgb}{0.79, 0.53, 0.38}
\definecolor{myred}{rgb}{0.7, 0.36, 0.37}
\definecolor{mygreen}{rgb}{0.37, 0.61, 0.42}
\definecolor{mymagenta}{rgb}{0.52, 0.47, 0.66}
\definecolor{myyellow}{rgb}{1.0, 0.75, 0.0}
\definecolor{mycyan}{rgb}{0.0, 0.72, 0.92}
\definecolor{codegreen}{rgb}{0,0.6,0}

\begin{tikzpicture}
\pgfplotsset{compat=1.5}
\begin{axis}[scatter/classes={
  a={mark=*,myyellow, scale=0.8},
  b={mark=*,myorange, scale=0.8},
  c={mark=*,myred, scale=0.8},
  d={mark=*,mygreen, scale=0.8},
  e={mark=*,myblue, scale=0.8},
  f={mark=*,mycyan, scale=0.8}},
  width=1\columnwidth,
  height=50mm,
  grid=both,
  legend columns=-1,
  minor x tick num=1,
  minor y tick num=1,
  set layers, 
  mark layer=axis tick labels,
  xlabel=Latency (ns),
  ylabel=Test Error Rate (\%),
  tick label style={font=\footnotesize},  
  legend style={at={(0.5,1.2)},anchor=north, font=\footnotesize},
]
\addlegendimage{empty legend}

\addplot [scatter, only marks, scatter src=explicit symbolic]coordinates {
    (2.476, 30.27) [a]
    (4.797, 28.6) [a]
    (6.38, 28.22) [a]
    (8.9, 28.14) [a]
};
\addplot [scatter, only marks, scatter src=explicit symbolic] coordinates {
    (2.624, 29.09) [b]
    (4.659, 27.85) [b]
    (7.456, 27.61) [b]
    (11.285, 27.7) [b]
};
\addplot [scatter, only marks, scatter src=explicit symbolic] 
coordinates {
    (2.784, 28.21) [c]
    (4.884, 27.9) [c]
    (6.692, 27.57) [c]
    (11.63, 27.68) [c]
};
\addplot [scatter, only marks, scatter src=explicit symbolic] 
coordinates {
    (3.08, 28.58) [d]
    (4.854, 27.73) [d]
    (7.092, 27.58) [d]
    (9.755, 27.54) [d]
};
\addplot [scatter, only marks, scatter src=explicit symbolic] 
coordinates {
    (2.724, 28.36) [e]
    (4.716, 27.99) [e]
    (7.644, 27.63) [e]
    (10.44, 27.59) [e]
};
\addplot [scatter, only marks, scatter src=explicit symbolic] 
coordinates {
    (2.684, 28.64) [f]
    (4.647, 27.71) [f]
    (7.248, 27.67) [f]
    (9.4, 27.54) [f]
};
\addplot [black,thick] 
coordinates {
    (2.476, 30.27)
    (4.797, 30.27)
    (4.797, 28.6)
    (6.38, 28.6)
    (6.38, 28.22)
    (8.9, 28.22)
    (8.9, 28.14)
};

\addplot [red,thick] 
coordinates {
    (2.476, 30.27)
    (2.624, 30.27)
    (2.624, 29.09)
    (2.684, 29.09)
    (2.684, 28.64)
    (2.724, 28.64)
    (2.724, 28.35)
    (2.784, 28.35)
    (2.784, 28.19)
    (4.62, 28.19)
    (4.62, 27.69)
    (6.692, 27.69)
    (6.692, 27.57)
    (9.4, 27.57)
    (9.4, 27.54)
};

\addplot [black, densely dashed, smooth,<-]
coordinates {
    (4.797, 28.6)
    (3.7605, 28.73)
    (2.724, 28.6)
}
node[
  text=black,
  font=\tiny,
  above,
  yshift=-0.1cm] at (1215,28.6) [pos=0.5,font=\tiny]{$\mathbf{1.76\times}$};;

\addplot [black, densely dashed, smooth,<-]
coordinates {
    (6.38, 28.22)
    (4.582, 28.32)
    (2.784, 28.22)
}
node[
  text=black,
  font=\tiny,
  above,
  yshift=-0.1cm] at (1215,28.6) [pos=0.5,font=\tiny]{$\mathbf{2.29\times}$};;

\addplot [black, densely dashed,<-]
coordinates {
    (8.9, 28.14)
    (6.76, 28.04)
    (4.62, 28.14)
}
node[
  text=black,
  font=\tiny,
  below,
  yshift=0.08cm] at (1215,28.6) [pos=0.5,font=\tiny]{$\mathbf{1.92\times}$};;

\draw[black,very thick,<-] (250,230) -- (530,230) node [fill=gray!30!white,pos=1.5, rounded corners=2pt, inner sep=2pt]
{Baseline Pareto Frontier};

\draw[black,very thick,<-] (640,10) -- (710,90) node [fill=gray!30!white,pos=1.3, rounded corners=2pt, inner sep=2pt]
{Our Pareto Frontier};

\addlegendentry{\hspace{-.0cm}\textbf{Degree}}
\addlegendentry{1}
\addlegendentry{2}
\addlegendentry{3}
\addlegendentry{4}
\addlegendentry{5}
\addlegendentry{6}
\end{axis}
\end{tikzpicture}}
	\caption{
            Analysis of the trade-off between test error rate and latency for four models with varying numbers of layers across six different polynomial degrees. Our improved Pareto frontier is depicted by (\ref{pgfplots:label1}).
        }
	\label{fig:g3}
\end{figure}
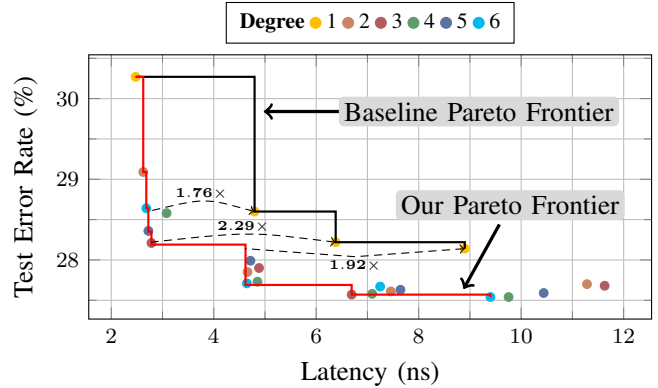

\subsection{Comparison with prior work}

\begin{table*}[htbp]
\caption{Evaluation of PolyLUT on multiple datasets. Bold indicates best in class.}
\begin{center}
\renewcommand{\arraystretch}{1.5} 
\begin{tabular}{clrrrrrrr}

&&\textbf{Accuracy}&\textbf{LUT}&\textbf{FF}&\textbf{DSP}&\textbf{BRAM}&\textbf{$\text{F}_\text{max}$ (MHz)}&\textbf{Latency (ns)}\\
\cellcolor[gray]{0.9}&\textbf{PolyLUT (NID Lite)}&\textbf{92\%}&\textbf{3336}&686&\textbf{0}&\textbf{0}&\textbf{529}&\cellcolor[gray]{0.9}\textbf{9}\\
\cline{2-9}
\cellcolor[gray]{0.9}&\textbf{LogicNets}\cite{logicnets}$^{\mathrm{a}}$&91\%&15949&1274&\textbf{0}&\textbf{0}&471&\cellcolor[gray]{0.9}13\\
\cline{2-9}
\multirow{-3}{*}{\cellcolor[gray]{.9}\textbf{UNSW-NB15}} &\textbf{Murovic \textit{et al.}}\cite{murovic}&\textbf{92\%}&17990&\textbf{0}&\textbf{0}&\textbf{0}&55&\cellcolor[gray]{0.9}18\\
\hline
\hline
\cellcolor[gray]{0.9}&\textbf{PolyLUT (HDR)}&\textbf{96\%}&\textbf{70673}&\textbf{4681}&\textbf{0}&\textbf{0}&\textbf{378}&\cellcolor[gray]{0.9}\textbf{16}\\
\cline{2-9}
\cellcolor[gray]{0.9}&\textbf{FINN}\cite{finn}&\textbf{96\%}&91131&-&\textbf{0}&5&200&\cellcolor[gray]{0.9}310\\
\cline{2-9}
\multirow{-3}{*}{\cellcolor[gray]{.9}\textbf{MNIST}} &\textbf{\texttt{hls4ml}}\cite{hls4ml}&95\%&260092&165513&\textbf{0}&\textbf{0}&200&\cellcolor[gray]{0.9}190\\
\hline\hline
\cellcolor[gray]{0.9}&\textbf{PolyLUT (JSC-M Lite)}&72\%&\textbf{12436}&\textbf{773}&\textbf{0}&\textbf{0}&\textbf{646}&\cellcolor[gray]{0.9}\textbf{5}\\
\cline{2-9}
\cellcolor[gray]{0.9}&\textbf{PolyLUT (JSC-XL)}&75\%&236541&2775&\textbf{0}&\textbf{0}&235&\cellcolor[gray]{0.9}21\\
\cline{2-9}
\cellcolor[gray]{0.9}&\textbf{LogicNets}\cite{logicnets}$^{\mathrm{a}}$&72\%&37931&810&\textbf{0}&\textbf{0}&427&\cellcolor[gray]{0.9}13\\
\cline{2-9}
\cellcolor[gray]{0.9}&\textbf{Duarte \textit{et al.}}\cite{duarte}&75\%&\multicolumn{2}{c}{88797$^{\mathrm{b}}$}&954&\textbf{0}&200&\cellcolor[gray]{0.9}75\\
\cline{2-9}
\multirow{-5}{*}{\cellcolor[gray]{.9}\textbf{Jet substructure tagging}} &\textbf{Fahim \textit{et al.}}\cite{fahim}&\textbf{76\%}&63251&4394&38&\textbf{0}&200&\cellcolor[gray]{0.9}45\\

\multicolumn{9}{l}{$^{\mathrm{a}}$New results can be found on the LogicNets GitHub page.}\\
\multicolumn{9}{l}{$^{\mathrm{b}}$Paper reports ``LUT+FF".}\\
\end{tabular}
\renewcommand{\arraystretch}{1.5}
\label{table:evaluation}
\end{center}
\vspace{-0.1cm}
\end{table*}

\vspace{0.2cm}
\subsubsection{Network intrusion detection}
For the network intrusion task, we evaluate our work against LogicNets~\cite{logicnets} and Murovic \textit{et al.}~\cite{murovic}. The results are summarized in Table~\ref{table:evaluation}.

For this task, we use the NID Lite architecture from Table~\ref{table:networks} with a polynomial degree set to $4$. PolyLUT achieves higher accuracy compared to LogicNets and the same accuracy as Murovic \textit{et al.} In terms of latency, PolyLUT achieves a $1.44\times$ and a $2\times$ reduction compared to LogicNets and Murovic \textit{et al.}, respectively. Additionally, PolyLUT improves the LUT count, with a reduction of $4.78\times$ compared to LogicNets and $5.39\times$ compared to Murovic \textit{et al.}.

\subsubsection{Handwritten digit recognition}
To evaluate the performance of PolyLUT on the MNIST digit classification dataset, we compare it against FINN's SFC-max model~\cite{finn} which is a fully unfolded implementation, and against the ternary neural network (TNN) used in \texttt{hls4ml}~\cite{hls4ml}. The results of the evaluation are shown in Table~\ref{table:evaluation}. 

We use the HDR architecture from Table~\ref{table:networks} for this task and train the network with a polynomial degree of $4$. PolyLUT achieves higher accuracy compared to \texttt{hls4ml} and the same accuracy as FINN. Notably, PolyLUT reaches a significant latency reduction of $19.38\times$ against FINN and a latency reduction of $11.88\times$ against \texttt{hls4ml}. Additionally, against FINN, PolyLUT has no block random-access memory (BRAM) utilization, and it reduces the LUT utilization by $1.28\times$. Against \texttt{hls4ml}, PolyLUT reduces the LUT utilization by $3.68\times$.

We achieve high latency reductions without compromising accuracy through the effective function complexity handling in our methodology and through the minimization of exposed datapaths, which are susceptible to causing bottlenecks.

\subsubsection{Jet substructure tagging}
To evaluate our method on the jet substructure tagging task, we conducted a comparative analysis against three existing approaches: LogicNets~\cite{logicnets}, the work presented by Duarte \textit{et al.}~\cite{duarte}, and the work presented by Fahim \textit{et al.}~\cite{fahim}. Our work demonstrates superior performance in terms of latency, surpassing all previous approaches. This highlights the effectiveness of our methodology in enabling applications that demand ultra-fast processing.

For our experiments, we utilized two architectures: JSC-M Lite and JSC-XL, with polynomial degrees set to $6$ and $4$, respectively, as outlined in Table~\ref{table:networks}. Table~\ref{table:evaluation} provides a comprehensive comparison of our work. 

In comparison to LogicNets, our JSC-M Lite architecture achieves the same accuracy while reducing the LUT count by a factor of $3.05\times$ and the latency by a factor of $2.6\times$. Compared to Duarte \textit{et al.}, our JSC-XL architecture maintains comparable accuracy while achieving a latency reduction of $3.57\times$. In comparison to Fahim \textit{et al.}, due to their higher precision (fixed-point), our method exhibits a decrease in accuracy of $1$ percentage point, however, we achieve an improvement in latency by a factor of $2.14\times$. 

In contrast to Duarte \textit{et al.} and Fahim \textit{et al.}, our approach does not use DSP blocks at all, however it can result in a large LUT count in specific cases where both a high neuron fan-in and high precision is required to reach a given accuracy. This situation can be seen in JSC-XL. 

\section{Conclusion and Further Work}
In this work, we introduced PolyLUT, a novel DNN-hardware co-design methodology designed to meet the stringent demands of on-edge applications, specifically in terms of ultra-low latency and minimal area requirements. We proposed mapping sparse and quantized polynomial neural networks to netlists of LUTs, by training multivariate polynomials instead of linear functions. Thus, we leveraged the capabilities of a LUT to represent any function and we maintained accuracy on shallower neural network models. 

We have demonstrated the effectiveness of our approach on three different datasets: network intrusion detection, handwritten digit classification, and jet substructure tagging. Compared to prior works, for similar accuracies, our method achieves significant latency improvements in these tasks, with reductions of up to $2\times$, $19.38\times$, and $3.57\times$, respectively.

While the number of training parameters scales polynomially for fixed D, the L-LUT size still scales exponentially, limiting network architectures with few inputs, each of low precision.

An evident extension of our work would be to incorporate neural architecture search, allowing for the exploration of optimal architectures and hyperparameters such as bit-width, fan-in, and degree; those in Table~\ref{table:networks} were obtained by hand. This approach could lead to architectures that offer improved performance and efficiency.

\clearpage
\bibliographystyle{IEEEtran}
\begingroup
\raggedright
\bibliography{bibs}
\endgroup

\end{document}